\documentclass[letterpaper,twocolumn,10pt]{article}
\usepackage{usenix-2020-09}

\usepackage{tikz}
\usepackage{amsmath}
\usepackage{pifont}
\usepackage{graphicx}
\usepackage{caption}
\usepackage{subcaption}
\usepackage{enumitem}
\usepackage{booktabs}
\usepackage{makecell}

\usepackage{hyperref}
\usepackage{cleveref}

\usepackage{listings}
\usepackage{color}

\definecolor{dkgreen}{rgb}{0,0.6,0}
\definecolor{gray}{rgb}{0.5,0.5,0.5}
\definecolor{mauve}{rgb}{0.58,0,0.82}
\usepackage{xcolor}
\usepackage{graphicx}
\usepackage{verbatim}
\usepackage{xspace}

\newcommand{\paragraphb}[1]{\vspace{1mm}\noindent\textbf{#1}\quad}

\newif\ifflow\flowfalse

\ifflow
\newcommand{\yukai}[1]{\textcolor{blue}{Yukai: {#1}}}
\newcommand{\yibo}[1]{\textcolor{red}{Yibo: {#1}}}

\newcommand{\TODO}[1]{\vskip1ex\noindent\colorbox{yellow}{\parbox{\columnwidth}{\textbf{TODO:}{#1}}}}
\else
\newcommand{\yukai}[1]{}
\newcommand{\yibo}[1]{}

\newcommand{\TODO}[1]{}
\fi

\newif\ifanon\anonfalse

\ifanon
\newcommand{\bytedance}{Corp X\xspace}
\newcommand{\belong}{XYExpress\xspace}
\newcommand{\beshort}{XE\xspace}
\newcommand{\bemalloc}{XYmalloc\xspace}
\else
\newcommand{\bytedance}{ByteDance\xspace}
\newcommand{\belong}{\textit{ByteExpress}\xspace}
\newcommand{\beshort}{\textit{BE}\xspace}
\newcommand{\bemalloc}{\textit{bemalloc}\xspace}
\fi

\newcommand{\etal}{\textit{et al.}\xspace}

\renewcommand{\paragraphb}[1]{\vspace{0.5mm}\noindent\textbf{#1}\quad}
\makeatletter
\renewcommand{\section}{\@startsection{section}{1}{\z@}
  {-2.4ex plus -0.6ex minus -0.2ex}{1.1ex plus 0.2ex}
  {\normalfont\large\bfseries\raggedright}}
\renewcommand{\subsection}{\@startsection{subsection}{2}{\z@}
  {-1.8ex plus -0.5ex minus -0.2ex}{0.8ex plus 0.2ex}
  {\normalfont\large\bfseries\raggedright}}
\renewcommand{\subsubsection}{\@startsection{subsubsection}{3}{\z@}
  {-1.5ex plus -0.4ex minus -0.2ex}{0.6ex plus 0.2ex}
  {\normalfont\normalsize\bfseries\raggedright}}
\renewcommand{\@maketitle}{\newpage
  \vbox{\vskip 4pt
    \centering
    {\Large\bfseries\@title\par}
    \vskip 7pt
    {\large\normalfont\lineskip 0.5em
      \begin{tabular}[t]{c}\@author\end{tabular}\par}
    \vskip 7pt}}
\makeatother

\begin{document}
\raggedbottom
\date{}
\title{\Large \bf Deploying RDMA for Distributed Inference and Training at Scale}
\author{
\normalsize\normalfont Yibo Huang$^{\dag,2,\S}$\thanks{This work was mostly done while Yibo Huang was pursuing his Ph.D. at Fudan University and interning at ByteDance Inc.}\quad 
Yi Cui$^{2}$\quad Yukai Huang$^{\dag,2}$\quad Huaping Zhou$^{2}$\quad Jiancong Tong$^{2}$\\
\normalsize\normalfont Zhaoyu Wu$^{2}$\quad Gen Miao$^{2}$\quad Haoran Wei$^{2}$\quad Boyan Ding$^{2}$\quad  Zhuo Jiang$^{2}$ \\
\normalsize\normalfont  Jianxi Ye$^{2}$\quad Yibo Zhu$^{2}$\quad  Chuanxiong Guo$^{2}$\quad Yang Xu$^{\dag}$\quad Jie Wu$^{\dag}$\\
\small\normalfont \textit{$^{\dag}$Fudan University\qquad $^{2}$ByteDance Inc.\qquad $^{\S}$University of Michigan}
}
\maketitle

\begin{abstract}
RDMA can reduce communication overhead in distributed machine learning systems, but realizing these benefits in production requires more than replacing TCP. Variable-size messages, many concurrent connections, bursty incasts, and service failures place demands on software that hardware offload alone does not address. This paper reports our experience deploying \belong, a software communication layer over hardware RDMA, in \bytedance's parameter-server-based inference and training systems. We describe three lessons from the deployment: efficient RDMA requires buffer management that spans the application and transport; scalable communication requires coordinated management of connections, receive capacity, and outstanding data; and service availability requires recovery above the transport layer. Our design combines an RDMA-aware allocator, zero-copy message transfers, QP multiplexing, receive-credit flow control, a software send window, and RPC-assisted recovery. In a production inference workload, the system reduces mean and 99th-percentile model parameter fetch latency by 35.7\% and 43.7\%, respectively, and total CPU usage by 10.69\% relative to the widely deployed TCP stack. 
In the training workload, \belong reduces P99 model parameter push latency by up to 74.0\% relative to TCP, with no PFC pauses observed during the reported interval. 
\end{abstract}

\section{Introduction}\label{sec:intro}

Distributed inference and training underpin recommendation, search, and advertising services at \bytedance, including Toutiao and TikTok. In the parameter-server architecture used by these services, workers fetch model parameters from multiple servers; trainers also push updates back to those servers. These exchanges place communication on the critical path of both user-facing inference requests and iterative training. By January 2022, \bytedance's inference and training systems spanned hundreds of clusters and tens of thousands of computing nodes.

Faster computation increases the pressure on this communication path. GPU-based workers can generate substantial network demand~\cite{choquette2021nvidia,narayanan2021efficient,jiang2020unified}, while processing data through the kernel network stack consumes CPU cycles and introduces copying and scheduling overhead. RDMA offers hardware transport offload and direct access to registered memory. \bytedance had already deployed RDMA over Converged Ethernet (RoCE) in its data centers~\cite{jiang2020unified,peng2019generic}. The remaining challenge was to integrate that infrastructure into inference and training services without sacrificing scalability or availability.

This paper reports that integration experience. We present \belong (\beshort), a software communication layer built on hardware RDMA NICs (RNICs). Applications continue to use Archon, our existing RPC framework. Archon passes messages to \beshort through a socket-like interface, and \beshort issues verbs to the RNIC using Reliable Connection (RC) transport. The RNIC provides transport offload; \beshort manages buffers, schedules transfers, and bounds outstanding work; Archon handles request-level recovery. This division of responsibility is central to the design.

The workload makes the integration challenging in three ways. First, message sizes and buffer lifetimes vary with the model and request, making fixed receive buffers awkward to expose directly to applications. Second, fan-out requests and fan-in responses create many logical connections and bursts of concurrent transfers. Third, the completion of an inference request or training iteration can depend on its slowest sub-request. Hardware transport reliability alone cannot recover from every failed node, stalled connection, or unavailable service replica.

Our experience yields three main lessons.
\begin{itemize}[leftmargin=10pt,itemsep=2pt,topsep=2pt]
    \item \textit{Zero-copy communication is an interface and memory-management problem.} We build \bemalloc on the existing \textit{rpmalloc} allocator to make dynamically allocated application buffers RDMA-capable. The resulting ownership model lets the communication layer hand received buffers to application threads without copying. Batched work requests and completion-based receive notification then reduce engine overhead ($\S$\ref{sec:performance}).
    \item \textit{Scaling requires controlling distinct resources at distinct layers.} QP multiplexing reduces connection-state pressure, and thread domains avoid contention in shared driver structures. Receive credits protect posted receive capacity, whereas a separate per-engine send window bounds outstanding bytes during incasts. These mechanisms complement RNIC congestion control and fabric flow control; they address different bottlenecks ($\S$\ref{sec:scalability}).
    \item \textit{Reliable transport is only one part of service availability.} We retain RPC-level timeouts, retries, replica selection, and TCP fallback, and combine them with connection monitoring and replicated deployment. This allows recovery decisions to use service context that is unavailable to the RNIC ($\S$\ref{sec:availability}).
\end{itemize}

The contribution is the cross-layer co-design across these mechanisms and the operational lessons from deploying them. 
At the time of this study, the RDMA-based ML systems had supported production recommendation services for over two years. 
For a production recommendation inference workload, RDMA reduced average PS fetch latency by 35.7\%, P99 latency by 43.7\%, and total CPU usage by 10.69\% relative to the widely deployed TCP. 
In the training workload, RDMA with the software send window reduced P99 push latency by 74.0\% relative to TCP and eliminated observed PFC pauses during the reported measurement interval ($\S$\ref{sec:deployment}).

\section{Background}
\label{sec:background}

\subsection{Inference and Training Systems}
\label{sec:background:system}

\begin{figure}
    \centering
    \includegraphics[width=0.7\columnwidth]{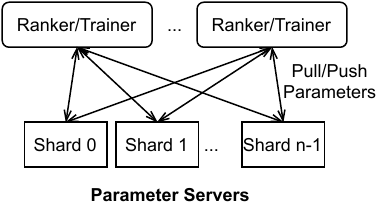}
    \caption{Parameter-server architecture: rankers fetch sharded model parameters for inference; trainers also push parameter updates.}
    \label{fig:system}

\end{figure}

At \bytedance, parameter-server-based systems~\cite{li2014scaling} support model training and personalized ranking for recommendation, search, and advertising. Figure~\ref{fig:system} shows the two main roles: workers and parameter servers (PSs). A worker is a \emph{ranker} during inference and a \emph{trainer} during training. Model parameters are partitioned into $n$ shards, for example by hashing, and stored across $n$ parameter servers.

For online inference, an application forwards a user request to a ranker. A request may require evaluating $k$ models. The ranker fetches the required parameters, evaluates the models, and returns the ranking results. Because the parameters can span multiple shards, one user request becomes multiple concurrent sub-requests to parameter servers. Each ranker also keeps multiple user requests outstanding through asynchronous network I/O. Request and response sizes vary with user attributes and model structure, ranging from kilobytes to megabytes. Prior work reports similarly variable message sizes~\cite{xue2019fast,dong2020eflops,naumov2020deep}.

For training, each trainer works on a model, loads training data, fetches parameters, performs forward and backward propagation, and pushes updates to parameter servers. Repeated parameter exchanges generate incasts at both workers and servers. Throughout the paper, \texttt{Fetch} denotes parameter retrieval; the synthetic experiment labels the corresponding operation \texttt{Pull}.

The measurements in this paper cover different workloads. In the training workload discussed in $\S$\ref{subsec:cc}, mean and P99 message sizes are 20\,KB and 50\,KB, respectively. The allocator benchmark instead uses synthetic large allocations of 64\,KB--1\,MB and small allocations of 16\,B--16\,KB ($\S$\ref{subsec:rdma-mem-mgr}). The production inference evaluation uses 300 parameter servers and 124 rankers; the training evaluation uses 124 parameter servers and 346 trainers ($\S$\ref{sec:deployment}). These examples characterize the evaluated cases, not a single distribution shared by all services.

\subsection{RDMA: A Primer}
\label{sec:background_rdma}

Remote Direct Memory Access (RDMA) provides low-latency, high-throughput communication using specialized NICs. RNICs offload transport processing and expose operations to user-space software, avoiding kernel-network-stack overhead on the data path. Software still allocates buffers, posts operations, and processes completions; RDMA does not eliminate application-side CPU work.

A queue pair (QP) is a communication endpoint with a send queue (SQ) and a receive queue (RQ). Applications access RNIC functionality through the \emph{verbs} interface, which supports resource creation, work-request submission, and completion polling. Completed operations are reported through a completion queue (CQ).

Two-sided messaging verbs, \texttt{SEND} and \texttt{RECV}, require participation from both endpoints: the receiver posts a receive buffer before the incoming send can be completed. One-sided memory verbs include \texttt{WRITE}, \texttt{READ}, and atomic operations; RDX proposes extending RDMA beyond memory access to remote code execution~\cite{huang2025remote}. 
They access registered remote memory without requiring the remote CPU to execute a matching operation for each transfer. The remote application must nevertheless arrange memory registration, access permissions, and buffer lifetimes.

\begin{table}[!th]
    \centering
    \begin{tabular} {lccc}
        \toprule
        & RC & UC & UD \\ \hline
        SEND/RECV & \ding{51} & \ding{51} & \ding{51} \\ \hline
        WRITE & \ding{51} & \ding{51} & \ding{55} \\ \hline
        READ/ATOMIC & \ding{51} & \ding{55} & \ding{55} \\ \hline
        Connection-oriented & \ding{51} & \ding{51} & \ding{55} \\ \hline
        Reliable & \ding{51} & \ding{55} & \ding{55} \\ \hline
        Maximum message size & 2GB & 2GB & MTU \\ 
        \bottomrule
    \end{tabular}

    \caption{Comparison of RDMA transports.}

    \label{tab:transport}
\end{table}

Table~\ref{tab:transport} summarizes three RDMA transport modes: Reliable Connection (RC), Unreliable Connection (UC), and Unreliable Datagram (UD). RC and UC are connection-oriented and support large messages; their supported memory operations differ. UD supports connectionless communication to multiple destinations, but each message is limited to the maximum transmission unit (MTU) and reliability must be supplied above the transport. \beshort uses RC to retain hardware reliability and large-message support while managing connection count in software.

In our deployment, RC runs on RoCE-capable hardware. DCQCN~\cite{zhu2015congestion} controls RNIC sending rates in response to congestion feedback, while Priority Flow Control (PFC) pauses traffic classes at the link layer to limit buffer overflow. RC reliability, congestion control, and PFC serve distinct purposes. Sections~\ref{subsec:engine} and~\ref{subsec:cc} explain the additional software controls used by \beshort.

\section{Deployment Requirements}\label{sec:motivation}

Our goal is to improve communication performance while retaining the scalability and availability expected of production inference and training services. The following workload properties determine our design choices.

\paragraphb{Variable messages and buffer lifetimes.}
Inference and training exchange messages whose sizes are not always known before a request is processed. Application threads may also retain or transfer ownership of a received buffer while computing a result. A fixed receive-buffer layout can complicate zero-copy integration in this setting. Registering a new buffer for every message avoids that layout constraint but adds registration overhead and increases the number of memory regions. We therefore need dynamic allocation from reusable, registered memory ($\S$\ref{subsec:rdma-mem-mgr}).

\paragraphb{Large transfers favor RC, but RC connections must be managed.}
UD-based designs such as FaSST and eRPC~\cite{kalia2016fasst,kalia2019datacenter} demonstrate the benefits of connectionless communication, particularly for small messages. Our workload also includes messages much larger than the MTU. Sending them over UD requires software fragmentation, reassembly, and reliability. Section~\ref{subsec:transport-choice} compares RC and UD throughput under several link-rate and MTU configurations. We choose RC for its hardware support for reliable, large-message transfers, then multiplex logical connections to limit QP-state pressure ($\S$\ref{subsec:qp-mgr}). This is a choice for our workload and deployment, not a general argument against UD.

\paragraphb{Receive capacity and fabric congestion are separate constraints.}
A sender can exhaust posted receive requests even when the network is uncongested. Conversely, a receiver with available buffers can still be reached through a congested network. We address the first problem with credits for operations that consume receive requests ($\S$\ref{subsec:engine}). We address the second by limiting outstanding bytes per engine, in addition to the existing DCQCN and PFC mechanisms ($\S$\ref{subsec:cc}). Keeping these controls separate makes their roles explicit: receive credits prevent receiver-not-ready (RNR) events; the send window limits the amount of data injected during concurrent transfers.

\paragraphb{Incast makes tail behavior a scaling concern.}
Many workers can contact the same parameter server, and responses from many servers can converge on a worker. Although a single flow's bandwidth-delay product may be small relative to switch buffering, the aggregate traffic from many concurrent flows can exceed that capacity. In our training deployment, hardware congestion control alone did not prevent sustained PFC pauses and poor request latency. This observation motivates a software bound on outstanding data, rather than assuming that the fabric can absorb every burst.

\paragraphb{Service recovery requires application context.}
An RNIC can provide reliable delivery on a functioning connection, but it cannot select another service replica or decide how to retry a failed RPC. Fan-out operations are particularly sensitive to slow or failed sub-requests. We therefore retain Archon's RPC abstraction and extend its existing availability mechanisms to cover RDMA connection failures and fallback to TCP ($\S$\ref{sec:availability}).

\section{System Overview}
\label{sec:overview}

\begin{figure}[t]
    \centering
    \includegraphics[width=\columnwidth]{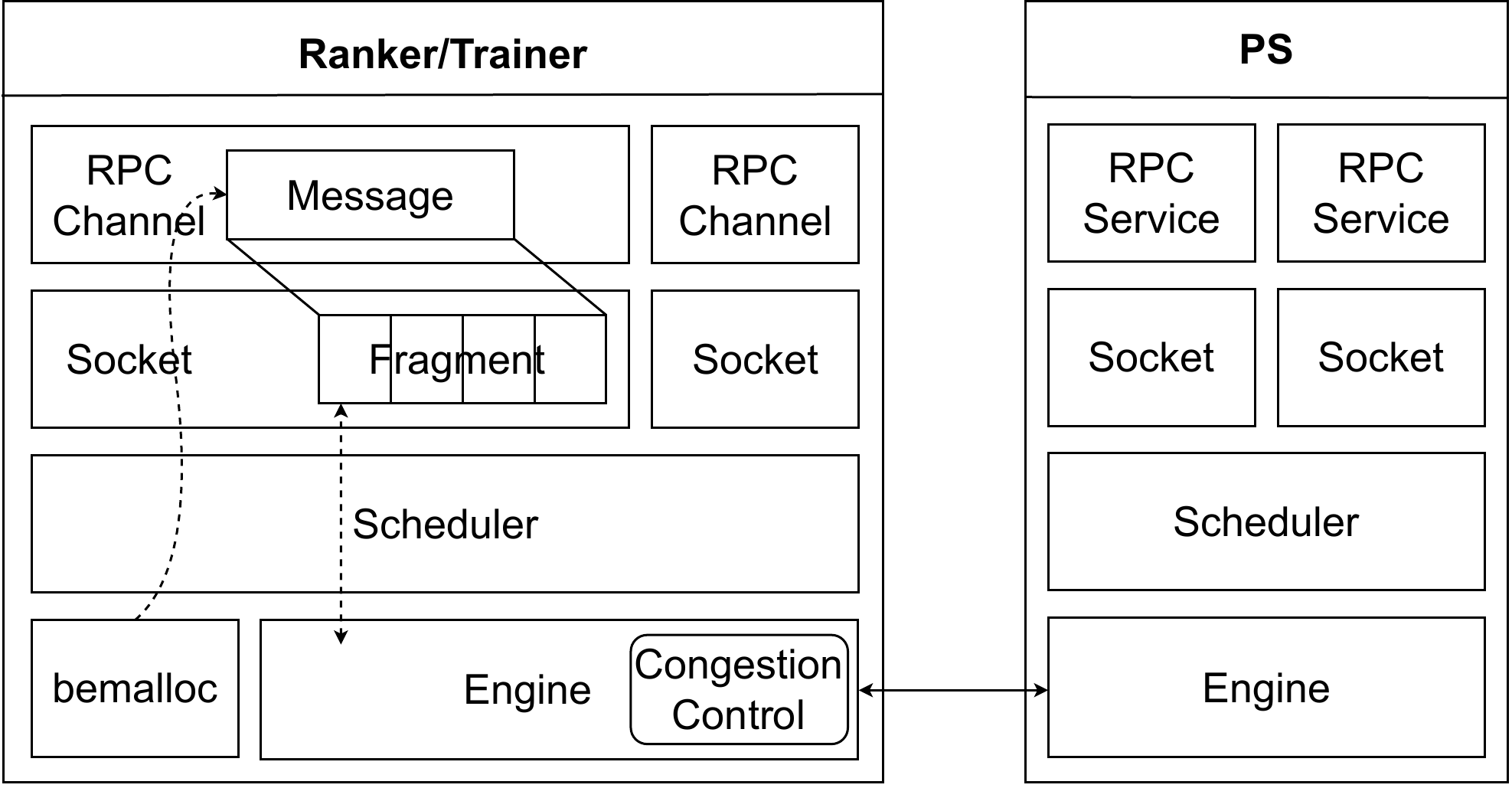}
    \caption{Communication stack: Archon provides RPCs; 
    \belong manages buffers and message transfers over hardware RDMA.}
    \label{fig:overview}
\end{figure}

Figure~\ref{fig:overview} shows the communication path between a ranker or trainer and a parameter server. Applications issue RPCs through Archon, our existing RPC framework. Archon uses \belong through socket-like endpoints, an abstraction also used by rack-scale PCIe interconnects~\cite{huang2022ultra}. \beshort translates message transfers into verbs operations on hardware RNICs. Applications therefore retain an RPC interface rather than directly managing verbs, QPs, or remote addresses.

\beshort separates logical connections from physical transport resources. A \emph{socket} represents a logical connection, such as an RPC channel. An \emph{engine} owns one RC QP and its associated transport state. Sockets with the same remote destination can share an engine, and a scheduler coordinates their transfers. This multiplexing reduces the number of QPs without requiring applications to share RPC channels explicitly.

The \bemalloc allocator supplies registered buffers to both the application and communication layers. On receipt, buffer ownership can pass to an application thread without a copy. For large outgoing messages, \beshort exchanges destination-buffer metadata and then transfers the payload using RDMA writes. It divides these transfers into fragments so that the scheduler can interleave sockets and enforce a per-engine send window.

The design uses three distinct forms of transport control. Receive credits track posted receive capacity for \texttt{SEND} and \texttt{WRITE\_WITH\_IMM}; a software send window limits outstanding bytes; and RNIC DCQCN controls the hardware sending rate. PFC remains enabled in the fabric as a link-level protection mechanism. Archon provides recovery above these layers through timeouts, retries, replica selection, and TCP fallback. The following sections describe how these components address performance, scalability, and availability.

\section{Building a High-Performance Communication Engine}
\label{sec:performance}

This section describes the data path, from transport selection and buffer allocation to message transfer and work-request processing. The key requirement is to preserve the benefits of hardware RDMA while supporting variable-size messages and application-controlled buffer lifetimes.

\subsection{Choosing RC for Large-Message IO}
\label{subsec:transport-choice}

\beshort uses RC to retain hardware support for reliable, large-message transfers. UD avoids maintaining a dedicated transport connection to each peer, but its MTU-sized messages require additional software processing for larger payloads. The measurements in Table~\ref{tab:uc-ud-cmp} illustrate how this tradeoff changes with link rate and MTU.

\begin{table}[t]
    \centering
    \setlength{\tabcolsep}{5pt}
    \begin{tabular}{cccc}
    \toprule
    Link rate & MTU & RC (Gbps) & UD (Gbps) \\ \hline
    25G & 1024B & 23.15 & 22.97 \\ \hline
    100G & 1024B & 92.26 & 48.29 \\ \hline
    100G & 4096B & 97.03 & 95.74 \\ \hline
    200G & 4096B & 191.36 & 114.70 \\
    \bottomrule
    \end{tabular}

    \caption{RC and UD throughput measured with \texttt{ib\_send\_bw} at different link rates and MTUs.}

    \label{tab:uc-ud-cmp}
\end{table}

At 25\,Gbps with a 1,024-byte MTU, RC and UD both approach the link rate. At 100\,Gbps with the same MTU, RC reaches 92.26\,Gbps while UD reaches 48.29\,Gbps; increasing the MTU to 4,096 bytes brings both close to line rate. At 200\,Gbps with a 4,096-byte MTU, RC reaches 191.36\,Gbps and UD reaches 114.70\,Gbps. Thus, packet-processing rate can constrain throughput as link bandwidth grows, even though UD performs well in other configurations. These measurements support our choice of RC for the evaluated workload, rather than a general claim that UD is always bandwidth-limited. Section~\ref{subsec:qp-mgr} explains how we manage the connection-state cost of that choice.

\subsection{RDMA Memory Management}
\label{subsec:rdma-mem-mgr}

To support zero-copy transfers of large, variable-size messages, we need dynamic allocation of registered memory. Ideally, allocating and freeing RDMA buffers should be as straightforward as using a general-purpose allocator such as \texttt{jemalloc}~\cite{facebook2010jemalloc} or \texttt{tcmalloc}~\cite{lee2014feedback}. We implement this interface by extending an existing allocator with memory registration and memory-region tracking.

We build \bemalloc (ByteExpress malloc) on \textit{rpmalloc}~\cite{mattias2017rpmalloc}, chosen for its compact implementation (approximately 3,000 lines of C) and allocation performance. The underlying allocator obtains large blocks of size $B$, for example 4\,MB, from the operating system and subdivides them to serve allocation requests. It can return a block when no allocations remain in use. We use its block-allocation and block-release callbacks to invoke \texttt{ibv\_reg\_mr()} and \texttt{ibv\_dereg\_mr()}, respectively, and store memory-region information in the block metadata. The same approach can be applied to allocators that expose comparable callbacks.

\paragraphb{Allocation interface.}
\bemalloc hides registration details behind a standalone interface while retaining the performance of the underlying allocator. Figure~\ref{fig:example} shows a wrapper that exposes a global RDMA memory pool through static functions. After initialization with an RDMA device context, \texttt{Allocate()} returns registered memory, \texttt{GetMR()} retrieves its memory-region metadata in $O(1)$ time, and \texttt{Deallocate()} returns the allocation to the pool. The interface adds RDMA-capable memory management rather than changing \textit{rpmalloc}'s allocation algorithm.

\begin{figure}[t]
\begin{lstlisting}[language=C++]
// Init memory pool for a given device context
RdmaMemoryPool::Init(ibv_ctx);
// Allocate a block of RDMA registered memory
void* block = RdmaMemoryPool::Allocate(size);
// Get MR in O(1) time when needed
const struct ibv_mr* mr = RdmaMemoryPool::GetMR(block);
// free the block
RdmaMemoryPool::Deallocate(block);
\end{lstlisting}

\caption{\bemalloc interface: allocate registered memory, look up memory-region metadata, and release the allocation.}

\label{fig:example}
\end{figure}

Registering large blocks also limits memory-region proliferation. Registering each small allocation separately can increase pressure on RNIC caches that hold memory-region metadata~\cite{nvidia2021optimized,kalia2019datacenter,burke2021prism,tsai2017lite}. \bemalloc instead amortizes each registration across many allocations. It also inherits \textit{rpmalloc}'s thread-local and global caches, including support for freeing memory from a thread other than the allocating thread. This supports the buffer-ownership transfers used by our inference and training systems.

\begin{figure*}
     \centering
     \begin{subfigure}[t]{0.23\textwidth}
         \centering
         \includegraphics[width=\textwidth]{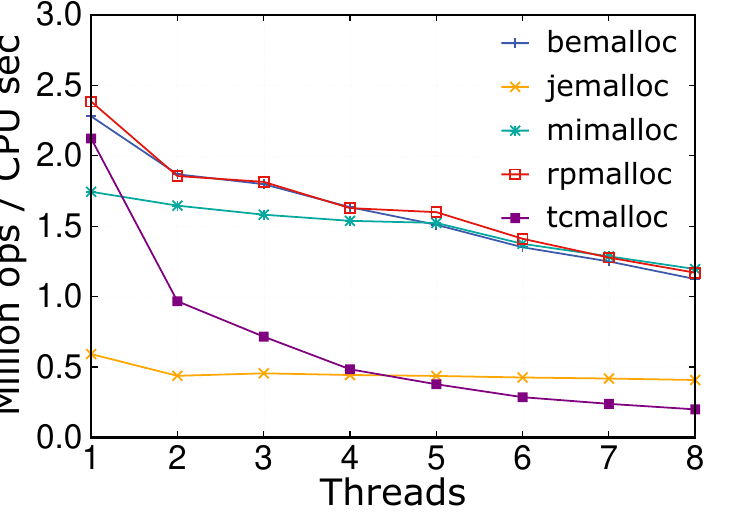}

         \caption{Large-block operation rate.}
         \label{fig:ops-large}
     \end{subfigure}
     \hfill
     \begin{subfigure}[t]{0.23\textwidth}
         \centering
         \includegraphics[width=\textwidth]{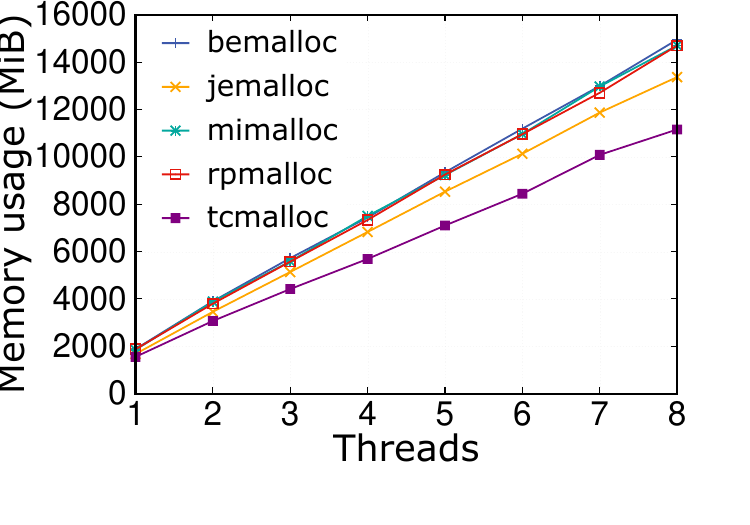}

         \caption{Large-block peak memory.}
         \label{fig:usage-large}
     \end{subfigure}
     \hfill
     \begin{subfigure}[t]{0.23\textwidth}
         \centering
         \includegraphics[width=\textwidth]{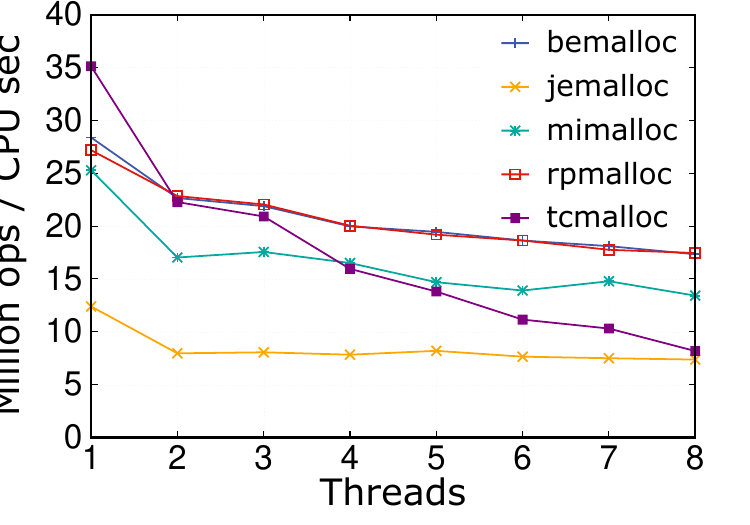}

         \caption{Small-block operation rate.}
         \label{fig:ops-small}
     \end{subfigure}
     \hfill
     \begin{subfigure}[t]{0.23\textwidth}
         \centering
         \includegraphics[width=\textwidth]{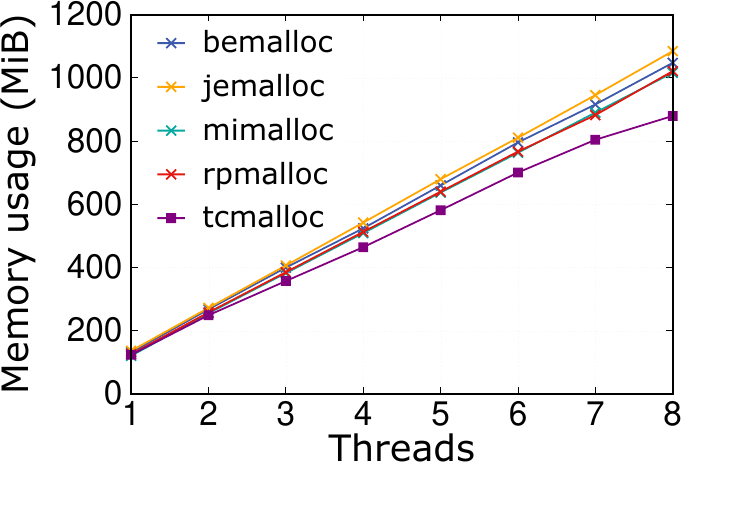}

         \caption{Small-block peak memory.}
         \label{fig:usage-small}
     \end{subfigure}

    \caption{Allocator operation rate and peak memory versus thread count. Large allocations span 64 KB–1 MB; small allocations span 16 B–16 KB.}
    \label{fig:compare-allocators}

\end{figure*}

\paragraphb{Registration cost and pool reuse.}
Registration remains necessary when the allocator acquires a new block, and deregistration is required when it releases one. Their cost is amortized across allocations from that block. Once the registered pool reaches a stable size, subsequent allocation and deallocation operations can be served from cached blocks without further registration. Thus, registration is largely removed from the steady-state allocation path; it is not free during pool growth or release.

\begin{figure}[t]
    \centering
    \includegraphics[width=0.8\columnwidth]{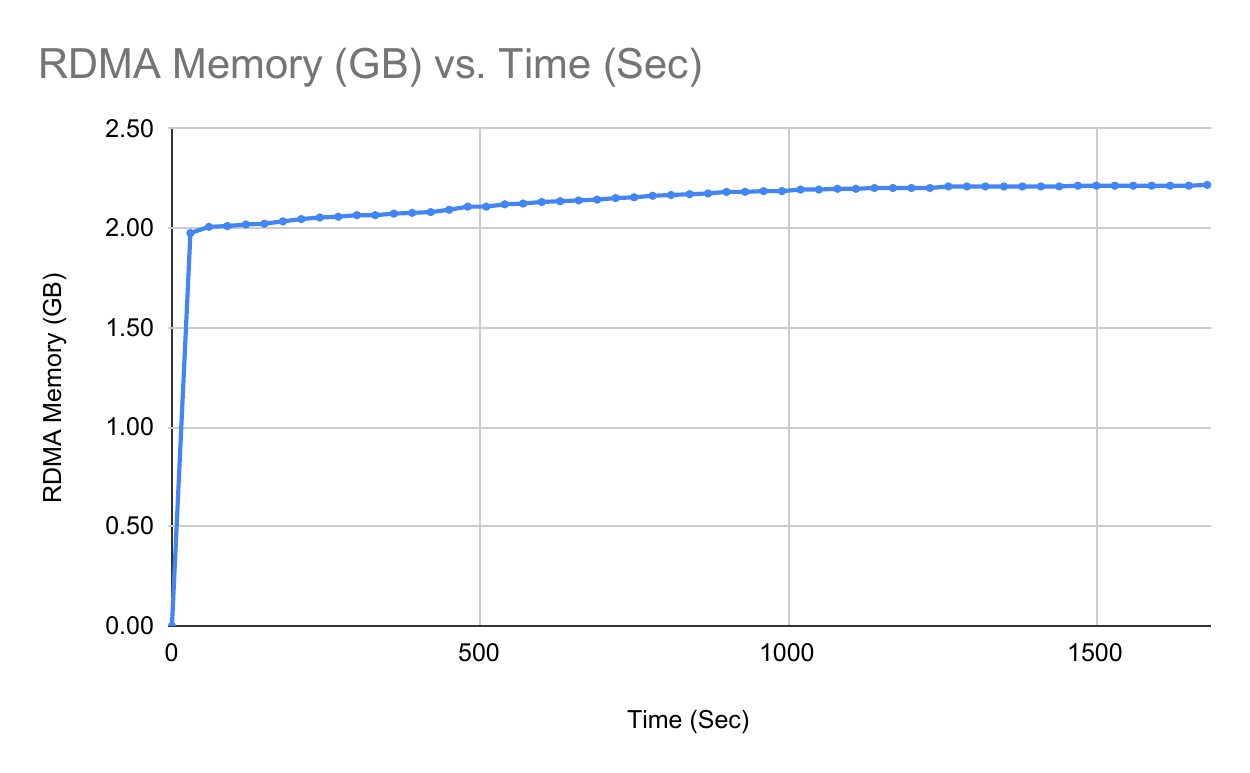}
    \vspace{-2mm}
    \caption{Registered RDMA memory footprint on one inference node, sampled every 30 s from startup and stabilizing near 2.2 GB.}
    \label{fig:ram}

\end{figure}

Figure~\ref{fig:ram} illustrates this behavior on one online inference node. Sampled every 30 seconds from startup, its registered-memory footprint reaches approximately 2\,GB in the first interval, then rises slowly to approximately 2.2\,GB and stabilizes. Allocation and deallocation continue after this point, but cached registered blocks serve those operations.

\textbf{Allocator evaluation.}
Figure~\ref{fig:compare-allocators} compares \bemalloc with general-purpose allocators. The large-allocation benchmark uses exponentially distributed sizes in the range 64\,KB--1\,MB; the small-allocation benchmark uses sizes in the range 16\,B--16\,KB with a linear size distribution. Both vary the thread count from one to eight and include approximately 10\% cross-thread operations. \bemalloc tracks \textit{rpmalloc} closely in operations per second and peak memory usage, indicating that the added RDMA bookkeeping has little overhead in these tests. They provide the highest operation rates among the evaluated allocators, at the cost of peak memory usage up to approximately 50\% higher than the alternatives.

\subsection{Message I/O}
\label{subsec:msg-io}

\subsubsection{Zero-Copy Buffer Ownership}

\bemalloc lets the application and communication layers share the same buffers. On receipt, the RNIC places message data directly into registered buffers allocated on demand. The communication layer then passes those buffers to the application, which can retain them, hand them to another thread, and free them after processing. This avoids coupling application buffer lifetimes to a fixed receive ring. Designs based on fixed receive buffers~\cite{dragojevic2014farm,kalia2014using,lu2017octopus,wei2018deconstructing} require a different ownership discipline to reuse those buffers safely.

On the send path, applications allocate network-bound payloads through \bemalloc, making them directly accessible to the RNIC. Other application allocations need not change. Here, \emph{zero-copy} means avoiding additional payload copies between application buffers and communication buffers; it does not imply that all model processing or serialization is copy-free.

\subsubsection{Transferring Large Messages}

\begin{figure}[t]
    \centering
    \includegraphics[width=0.9\columnwidth]{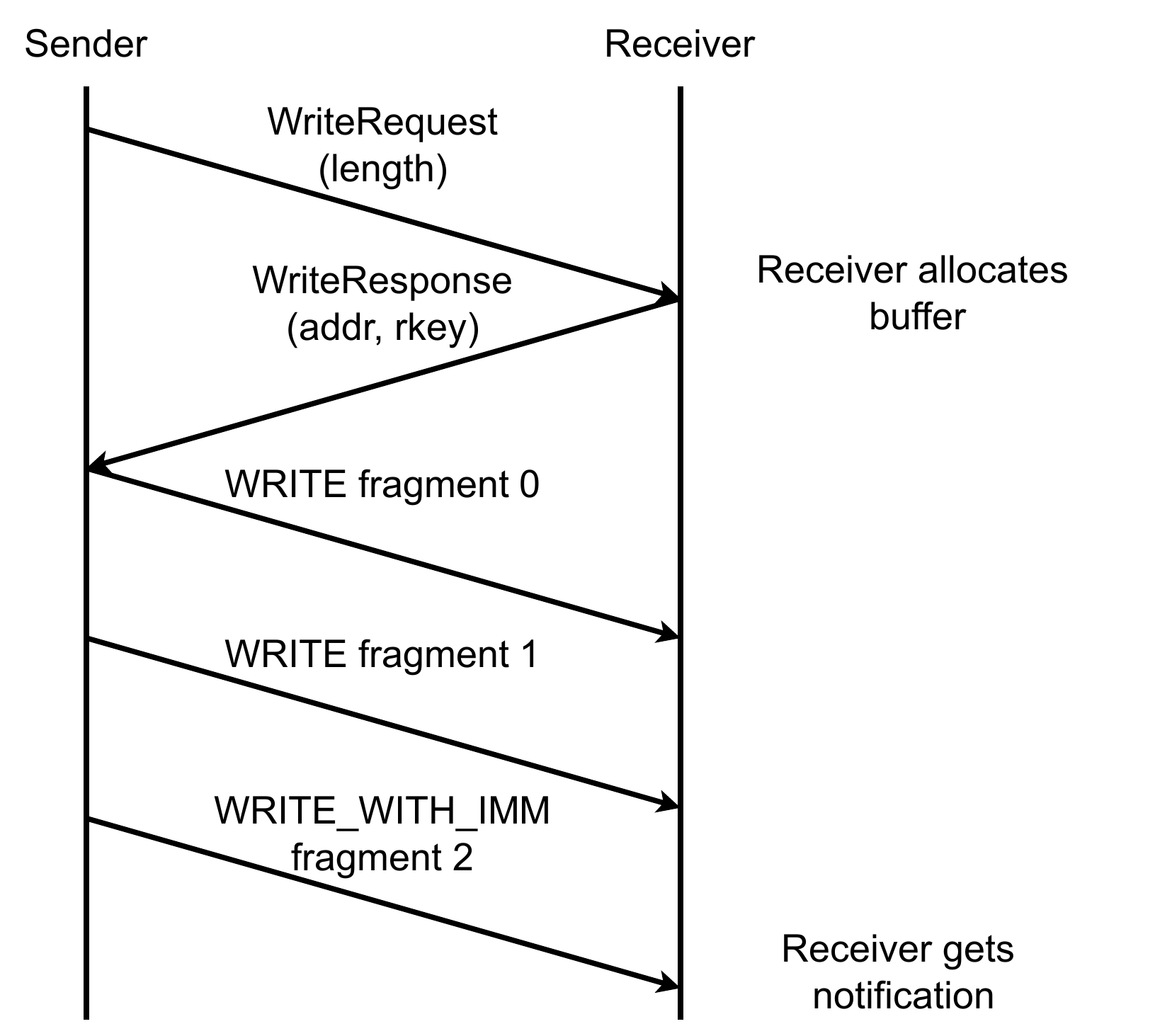}

    \caption{Large-message transfer: exchange destination-buffer metadata, write payload fragments, and notify the receiver with a final \texttt{WRITE\_WITH\_IMM}.}
    \label{fig:msg-io}

\end{figure}

\beshort uses one-sided \texttt{WRITE} operations for large payloads. Figure~\ref{fig:msg-io} shows the transfer sequence. The sender first obtains the destination address and remote memory key through a request/response exchange using \texttt{SEND}/\texttt{RECV}. The receiver allocates the destination buffer through \bemalloc. This exchange adds one round-trip time (RTT) before the payload transfer. If an application can supply valid remote-buffer metadata in advance, as with some statically allocated tensor buffers, \beshort can omit that exchange.

After exchanging metadata, \beshort writes the payload in fragments, with scheduling opportunities between fragments. For a destination address $p$ and fragment size $S$, successive fragments are written to $p$, $p+S$, and so on, producing a contiguous message at the receiver. Fragmentation lets the scheduler interleave sockets and enforce the send window described in $\S$\ref{subsec:cc}; the receiver does not need to copy fragments into a separate reassembly buffer.

\beshort sends the final fragment with \texttt{WRITE\_WITH\_IMM}, which carries 32 bits of immediate data and generates a receiver-side completion. The receiver uses this completion to detect that the message has arrived. By contrast, polling a flag embedded in a receive buffer~\cite{dragojevic2014farm,kalia2016design,kalia2014using,lim2014mica} requires a safe ordering relationship between the payload and the flag.

This distinction matters when memory regions use relaxed ordering. Relaxed ordering can permit incoming writes to reach memory out of order, so observing a flag alone is not sufficient evidence that the payload is visible. Completion semantics still ensure data visibility, including prior transfers~\cite{rdmacore_reg_mr}. Our notification scheme relies on those completion semantics for fragments issued on the same RC QP. It also lets an I/O thread poll a shared CQ instead of inspecting a separate flag for every outstanding message. The performance benefit of relaxed ordering is hardware-dependent; we do not isolate it in the evaluation.

\subsection{RDMA Engine}
\label{subsec:engine}

\begin{figure}[t]
    \centering
    \includegraphics[width=\columnwidth]{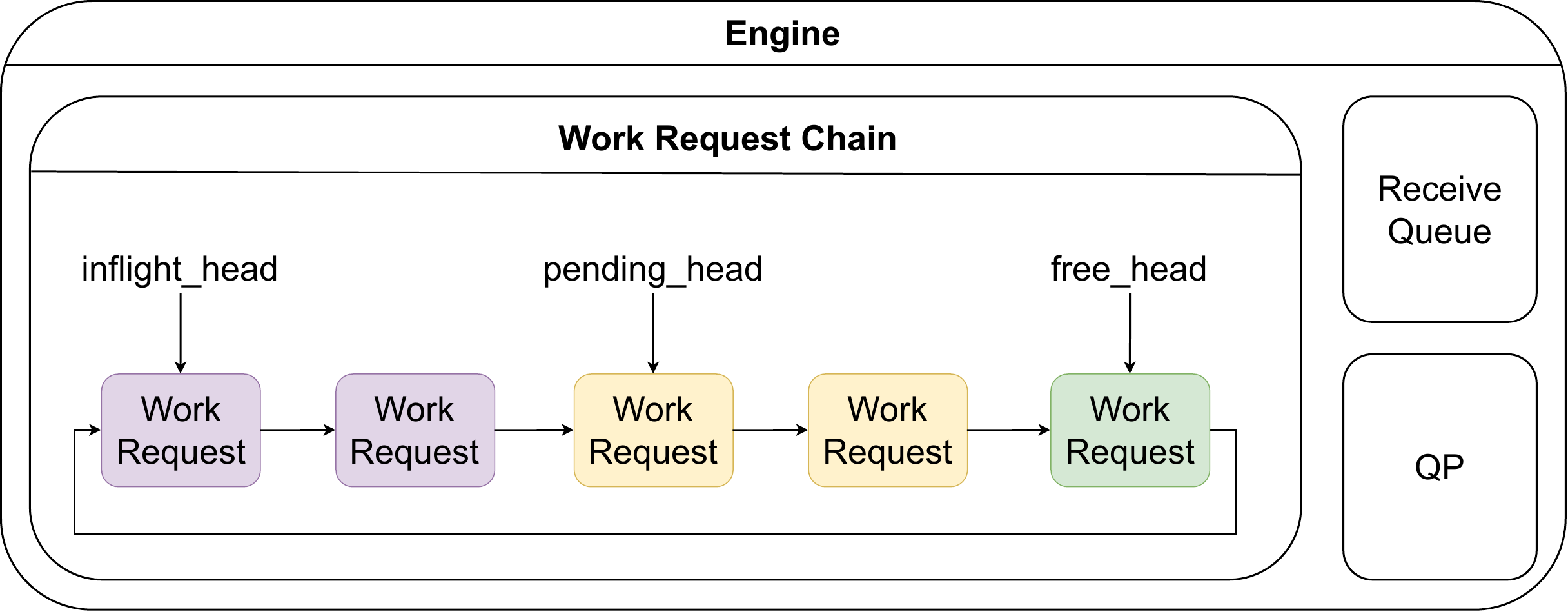}
    \caption{RDMA engine: one QP, receive resources, and a work-request chain partitioned into in-flight, pending, and free entries.}
    \label{fig:engine}
\end{figure}

Each \beshort engine owns one QP and its associated state, including a work-request chain and receive buffers (Figure~\ref{fig:engine}). It is the lowest software layer: it posts verbs operations and processes completions. Above it, sockets represent logical connections and a scheduler coordinates their I/O. We use \emph{engine} when referring to this software state and \emph{QP} when referring to the RDMA resource. Section~\ref{subsec:qp-mgr} describes socket multiplexing and scheduling.

\subsubsection{Work Request I/O}

The engine uses preallocated resources and batching to reduce data-path overhead. Its send-side work requests form a circular chain whose length equals the QP send-queue depth. Tracking occupied entries avoids dynamic work-request allocation and prevents the engine from posting more requests than the send queue can hold. We use relatively shallow queues, for example a depth of 32, provided that they allow enough outstanding work to sustain the target throughput.

\textbf{Work-request chain.} The chain has three contiguous segments: \emph{in flight}, \emph{pending}, and \emph{free}. In-flight requests have been posted to the RNIC and await completion; processing their completions frees the corresponding slots. Pending requests have been prepared but not posted. The engine submits them together with one \texttt{ibv\_post\_send()} call, using doorbell batching~\cite{kalia2016design}. Only the last request in a batch requests a completion notification, reducing completion-processing overhead. When free slots are available, the scheduler polls sockets for work; each selected socket prepares a request of at most one fragment. The engine tracks the segment boundaries in the circular chain.

\subsubsection{Flow Control}
\label{subsec:receive-flow-control}

\beshort uses \texttt{SEND}/\texttt{RECV} for control messages and small payloads. A \texttt{SEND} consumes a posted receive request at the peer; if none is available, the peer reports receiver not ready (RNR), causing retries and potentially a transport error. \texttt{WRITE\_WITH\_IMM} also consumes a receive request. We therefore track receive capacity explicitly. This flow control protects endpoint receive queues; it is separate from the network congestion control in $\S$\ref{subsec:cc}.

\textbf{Receive-credit flow control.} Inspired by receiver-driven transport~\cite{montazeri2018homa}, each sender maintains credits for available receive requests at its peer. The endpoints exchange the initial credit count during connection setup. Posting a \texttt{SEND} or \texttt{WRITE\_WITH\_IMM} consumes one credit; ordinary \texttt{WRITE} operations do not consume credits because they do not use receive requests. The sender waits when it lacks the required credit.

The receiver replenishes capacity in batches. After consuming $k$ receive requests, it posts $k$ replacements with one \texttt{ibv\_post\_recv()} call and sends a credit update to the peer. Batching reduces the frequency of posting calls and credit messages. We use $k=d/2$, where $d$ is the receive-queue depth. Choosing $k<d$ leaves capacity available while an update is in transit, although a sufficiently delayed update can still stall the sender. This scheme targets RNR avoidance with dedicated receive queues. The optional shared-receive-queue configuration trades this per-peer accounting for shared buffer capacity ($\S$\ref{subsec:per-thread-resources}).

\begin{figure*}
     \centering
     \begin{minipage}[t]{0.23\textwidth}
         \centering
         \includegraphics[width=\textwidth]{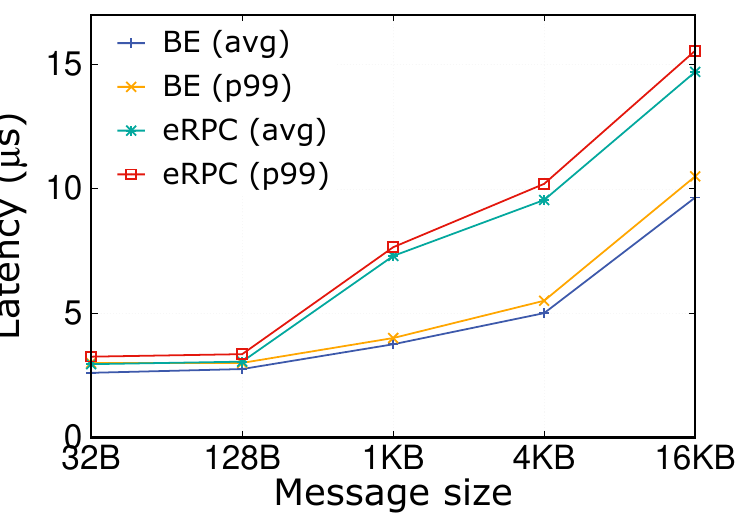}

         \caption{ByteExpress and eRPC latency with one message outstanding.}

         \label{fig:eval-msg-latency}
     \end{minipage}\hfill
     \begin{minipage}[t]{0.23\textwidth}
         \centering
         \includegraphics[width=\textwidth]{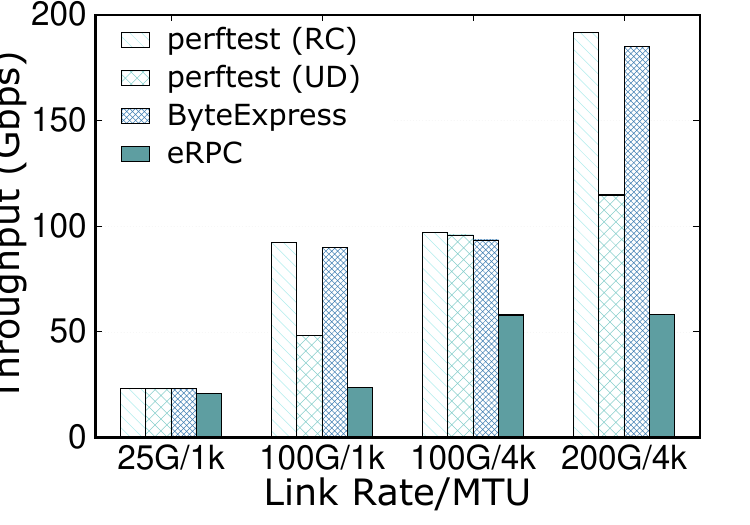}

         \caption{Peak throughput by link rate/MTU---ByteExpress vs. eRPC vs. RC/UD perftest.}

         \label{fig:eval-peak-tput}
     \end{minipage}\hfill
     \begin{minipage}[t]{0.23\textwidth}
        \centering
        \includegraphics[width=\textwidth]{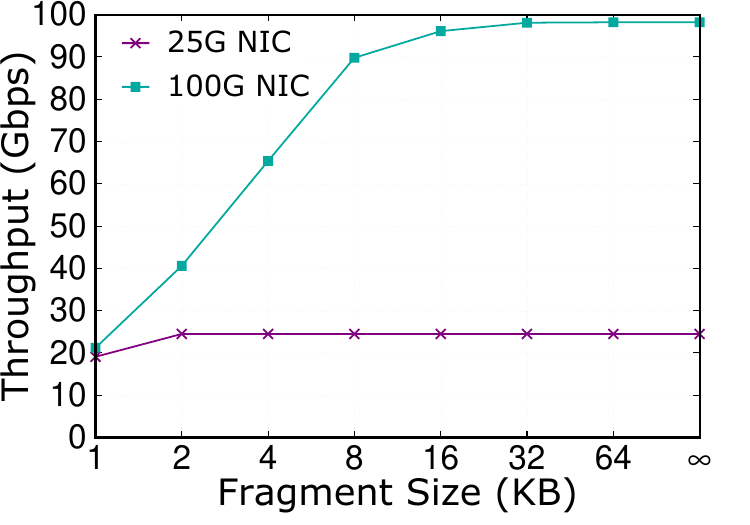}

        \caption{Throughput vs. fragment size on 25 and 100 Gbps links.}

        \label{fig:frag}
    \end{minipage}\hfill
    \begin{minipage}[t]{0.23\textwidth}
         \centering
         \includegraphics[width=\textwidth]{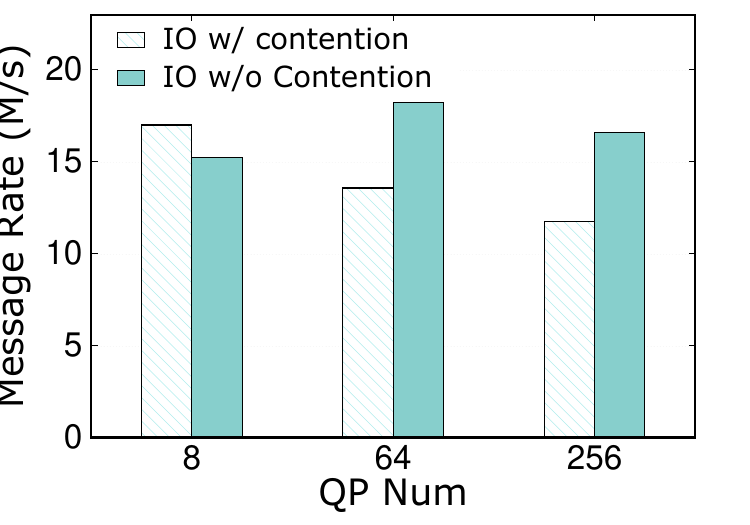}

         \caption{512-byte message rate vs. QP count in eight IO threads w/ or w/o contention.}

         \label{fig:qp}
     \end{minipage}
\end{figure*}

\subsection{Evaluation}
\label{subsec:perf-eval}

We evaluate \belong with synthetic workloads. Figure~\ref{fig:eval-msg-latency} compares mean and P99 message latency with eRPC~\cite{kalia2019datacenter} for message sizes from 32\,B to 16\,KB. The latency experiment permits one message to remain outstanding at a time.

Figure~\ref{fig:eval-peak-tput} compares peak throughput across four link-rate/MTU configurations: 25\,Gbps/1\,KB, 100\,Gbps/1\,KB, 100\,Gbps/4\,KB, and 200\,Gbps/4\,KB. It includes eRPC and RC/UD \texttt{perftest} baselines. The throughput setup uses one QP to send 1\,MB messages between two nodes, with a 32-byte response per message and sufficient concurrency to saturate the path.

\belong achieves latency and throughput comparable to or better than the plotted alternatives in these configurations. These results measure the complete communication path; they do not isolate the contribution of each optimization. The latency and throughput plots use different message-size settings and should be interpreted separately.

\section{Scaling Connections \& Concurrent PS Ops}
\label{sec:scalability}

Scaling the communication path requires managing both connection state and aggregate traffic. This section describes QP multiplexing, thread-aware resource management, and a software send window for incast workloads.

\subsection{QP Management}
\label{subsec:qp-mgr}

RNICs have limited cache space for QP context. When the active working set exceeds that capacity, the NIC must fetch missing context from host memory through DMA, increasing overhead. \beshort reduces this pressure by multiplexing logical connections onto QPs.

Archon keeps multiple requests outstanding through RPC channels, each with its own socket. With TCP, each socket has a separate transport connection. With \beshort, sockets directed to the same remote address can share one engine ($\S$\ref{subsec:engine}).

\textbf{Multiplexing sockets onto QPs.} \beshort maintains a connection table that maps remote addresses to engines. When an application calls \texttt{connect()}, \beshort attaches the socket to an existing engine for that address, if available. Otherwise, it creates an engine and reports connection success asynchronously after setup completes. This allows one QP to serve multiple logical connections to each remote address.

Sharing an engine introduces scheduling head-of-line (HoL) blocking: a large message from one socket can delay smaller messages from another. \beshort fragments messages so that the scheduler can interleave their work requests. Fragment size trades scheduling granularity against per-request overhead. Figure~\ref{fig:frag} shows that approximately 2\,KB fragments can saturate the tested 25\,Gbps link with one QP, whereas fragments larger than 8\,KB are needed for near-line-rate throughput on the tested 100\,Gbps link.

Each socket queues outgoing messages. When selected by the scheduler, it prepares a work request for at most one fragment by supplying that fragment's address and length; no payload copy is required. Under round-robin scheduling with $n$ active sockets and maximum fragment size $S$, at most $(n-1)S$ bytes from other sockets are selected between a socket's turns. This is a bound on scheduler service between turns, not on end-to-end latency: already-posted work, congestion, or loss recovery can still stall the shared QP. We use round-robin scheduling because sockets have comparable roles in the evaluated systems; other policies could prioritize latency-sensitive traffic.

\begin{figure*}[t]
     \centering
     \begin{minipage}[t]{0.33\textwidth}
         \centering
         \includegraphics[width=\textwidth]{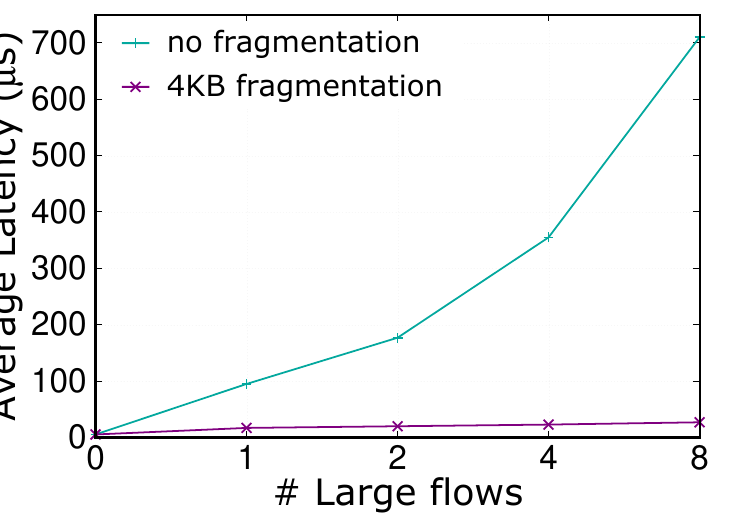}

         \caption{Small-flow latency under competing large flows, w/ or w/o 4KB frag.}

         \label{fig:small-flow-lat-for-frag}
     \end{minipage}
    \begin{minipage}[t]{0.65\textwidth}
         \centering
         \includegraphics[width=\textwidth]{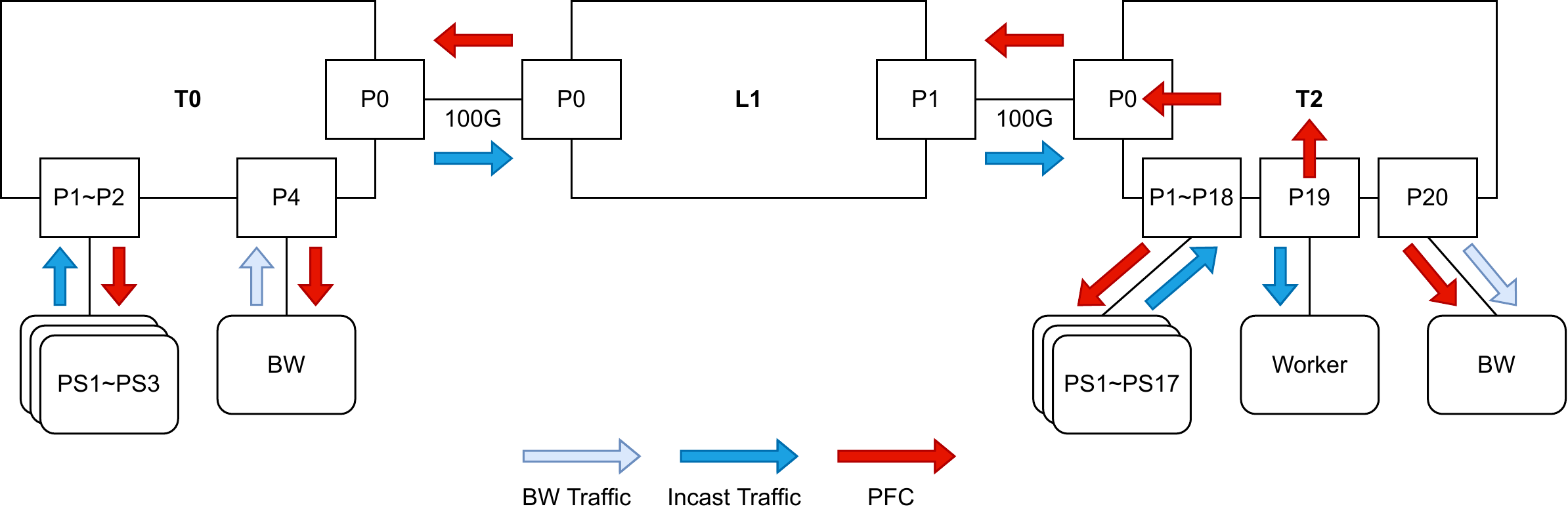}

         \caption{Synthetic incast topology showing workers, parameter servers, bandwidth-test flows, incast traffic, and PFC propagation.}

         \label{fig:simulation}
     \end{minipage}
\end{figure*}

Figure~\ref{fig:small-flow-lat-for-frag} adds large flows to an engine that already carries a small flow. With 4\,KB fragments, the small flow's latency still increases as contention grows, but remains substantially lower than without fragmentation. Fragmentation therefore mitigates message-scheduling HoL blocking; it does not eliminate blocking caused by transport ordering, packet loss, or congestion control.

\begin{figure}[t!]
    \centering
    \includegraphics[width=0.7\columnwidth]{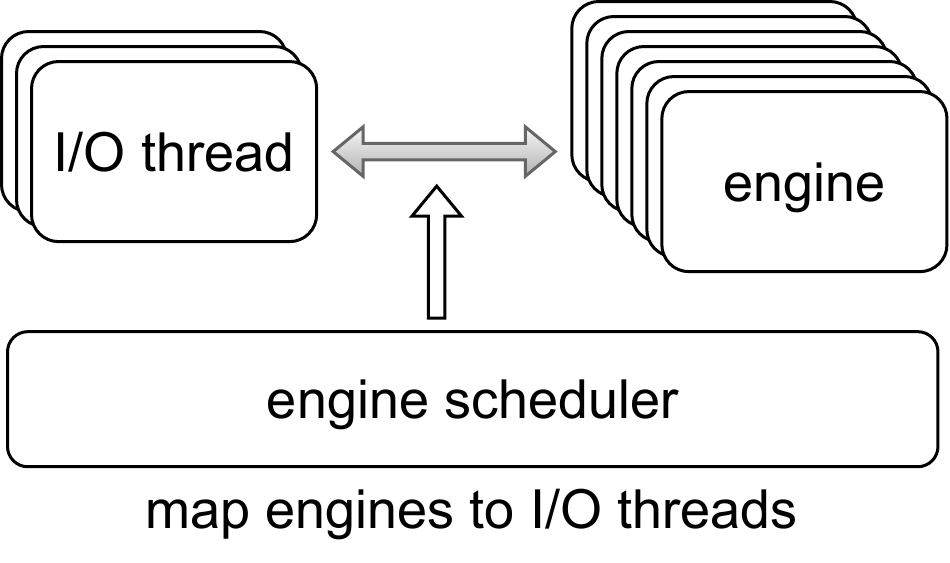}
    \caption{RDMA engine assignment---each engine (and its QPs) belongs to one I/O thread; one thread manages multiple engines.}
    \label{fig:mul-qp}
\end{figure}

\subsection{Avoiding Contention in Multithreaded I/O}
\label{subsec:multi-thread}

\subsubsection{Thread Ownership and Provider Resources}

Figure~\ref{fig:mul-qp} shows the mapping of engines to I/O threads. Each engine is bound to one thread, so operations on its data structures do not require locks between I/O threads. A thread can manage many engines. We assign engines in round-robin order, which is sufficient for the evaluated workload.

\textbf{Contention inside the RDMA provider.} Single-thread ownership of each engine did not eliminate all contention. In an experiment with eight threads and a varying number of QPs, the 512-byte message rate dropped sharply beyond 64 QPs (Figure~\ref{fig:qp}). Profiling showed substantial time in spin locks inside \texttt{ibv\_post\_send()}. QPs assigned to different application threads were sharing provider-side structures: the provider did not automatically inherit the application's ownership arrangement.

\textbf{Thread-aware resource allocation.} We use a separate thread domain (TD) for each I/O thread to communicate the ownership arrangement to the verbs provider. Thread domains permit provider optimizations for resources accessed by a single thread~\cite{rdmacore_alloc_td}. This requires the application to respect that ownership discipline. Separating threads into independent protection domains (PDs) would complicate cross-thread use of registered memory; instead, our configuration retains a common protection domain while associating resources with their respective thread domains.

\subsubsection{Sharing Queues Within an I/O Thread}
\label{subsec:per-thread-resources}

Single-thread ownership also lets engines share selected resources without introducing locks between I/O threads. Table~\ref{tab:resource} summarizes the allocation: QPs remain associated with remote addresses, while CQs and TDs are allocated per thread and registered memory belongs to a common PD. This reduces per-QP overhead, although QP-specific state still grows with connection count.

\begin{table}[t]
    \centering
    \begin{tabular} {ll} 
        \toprule
        \textbf{Resource} & \textbf{Allocation} \\ \midrule
        QP & One per remote address \\
        RQ & One per QP, without SRQ \\
        SRQ & One per thread, when enabled \\
        CQ & One per thread \\
        TD & One per thread \\
        MR & One per registered block \\
        PD & One globally \\ 
        \bottomrule
    \end{tabular}

    \caption{Allocation of RDMA resources in \beshort. Dedicated RQs and per-thread SRQs are alternative configurations.}

    \label{tab:resource}
\end{table}

\paragraphb{Completion queues.}
Each I/O thread maintains one CQ for the send and receive queues of its assigned QPs. It retrieves completion queue entries (CQEs) in batches with \texttt{ibv\_poll\_cq()} and uses each entry's QP number to identify the corresponding engine. This extends the engine's batched completion processing ($\S$\ref{subsec:engine}) across connections and avoids polling a separate CQ for every QP.

\paragraphb{Shared receive queues.}
Dedicated RQs reserve receive buffers for individual peers. As an alternative, \beshort can create one shared receive queue (SRQ) per I/O thread, allowing all engines assigned to that thread to draw from a common buffer pool instead of using dedicated RQs. This improves buffer utilization but changes the flow-control contract.

The credit scheme in $\S$\ref{subsec:receive-flow-control} tracks per-peer receive capacity and does not directly apply to a shared pool. Engines can be created and destroyed dynamically, so precise allocation of shared capacity requires additional coordination. In SRQ mode, we use larger queues and configure RNR timeouts and retry counts to reduce the impact of receive-buffer exhaustion. This mitigates RNR events but does not enforce the per-peer receive-capacity bound of the dedicated-queue credit scheme. The choice is therefore between explicit per-peer capacity accounting and more flexible sharing of receive buffers.

\subsection{Application-Level Congestion Control}
\label{subsec:cc}

Incast places a different limit on scale: many individually efficient senders can collectively overwhelm a receiver's network path. We supplement hardware congestion control with a software bound on outstanding bytes.

\textbf{Iterative incasts in training.} Both inference and training exhibit fan-in and fan-out, but repeated parameter synchronization in training can sustain concurrent transfers to the same destinations. In the training workload considered here, mean and P99 message sizes are 20\,KB and 50\,KB. Concurrent \texttt{Fetch}/\texttt{Push} requests can produce substantial incast congestion and frequent received PFC pause frames.

Our objective is to sustain concurrent transfers with low tail latency while reducing PFC activity. Absence of PFC pauses does not imply absence of congestion; it indicates that the measured workload did not trigger that protection mechanism. We use request P99 latency as a proxy for communication stalls, rather than as a direct measure of end-to-end training throughput.

Prior work has also identified congestion challenges in RDMA networks~\cite{kalia2019datacenter,zhu2015congestion,gao2018dcqcn+}. In our deployment, DCQCN alone did not react quickly enough to the observed incasts. Its rate-control dynamics motivate additional control over how much data software makes outstanding. Software transports such as eRPC~\cite{kalia2019datacenter} offer a different design point; we retain RC offload and add a software admission bound above it.

\textbf{A per-engine send window.} \beshort bounds the bytes in flight on each engine using a configurable send window, denoted \texttt{Cwnd} in the experiments. Together with fragmentation, this limits the amount of data an engine can make outstanding. DCQCN continues to regulate the RNIC's sending rate, and PFC remains enabled in the fabric. RC supplies transport reliability. These roles should not be conflated: the software window is an outstanding-data limit, DCQCN is a rate-control mechanism, and PFC is link-level backpressure.

Fragment and window sizes are chosen according to message sizes and incast scale. This configuration-based approach suits our training workloads with relatively stable traffic patterns, but it is not an adaptive congestion-control algorithm. A per-engine bound also does not impose a fixed bound on aggregate traffic as the number of active engines grows. We have not established the stability or fairness of the combined controls across arbitrary workloads.

\begin{figure}[t]
     \centering
     \begin{subfigure}[t]{0.235\textwidth}
         \centering
         \includegraphics[width=\textwidth]{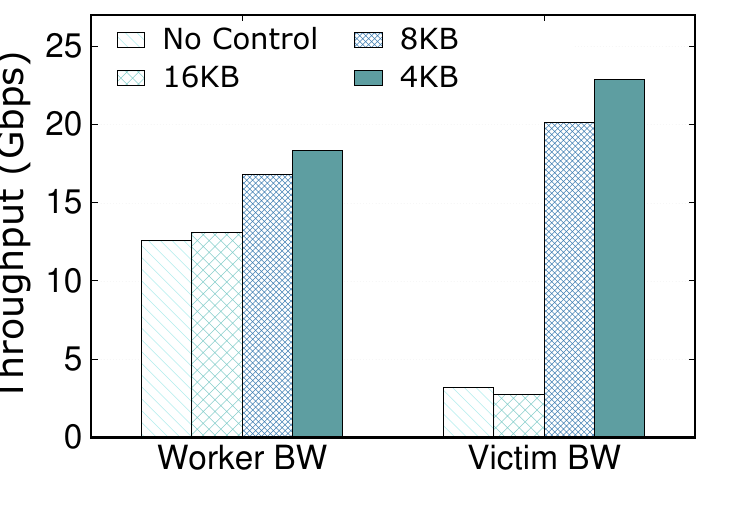}
         \vspace{-5mm}
         \caption{Worker and victim bandwidth.}
         \label{fig:simulation-bw}
     \end{subfigure}
     \hfill
     \begin{subfigure}[t]{0.235\textwidth}
         \centering
         \includegraphics[width=\textwidth]{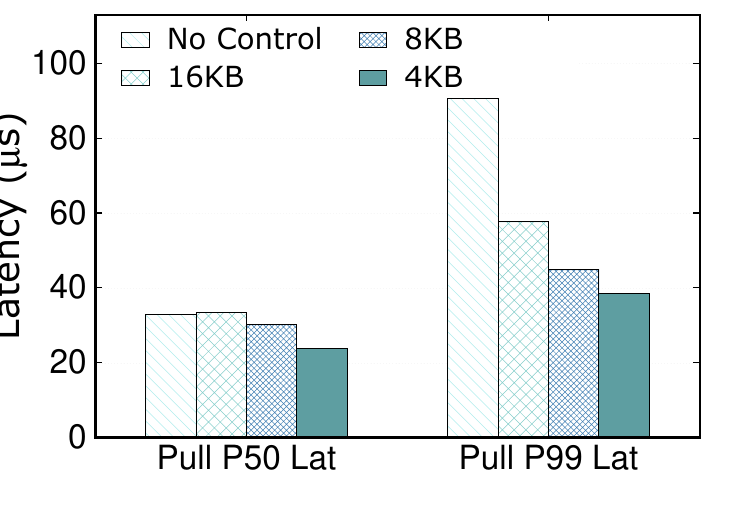}
         \vspace{-5mm}
         \caption{P50 and P99 pull latency.}
         \label{fig:simulation-lat}
     \end{subfigure}

    \caption{PS performance in the synthetic inference workloads with no software window or 4, 8, and 16 KB windows.}

    \label{fig:simulation-results}
\end{figure}

Figure~\ref{fig:simulation} shows the synthetic worker/parameter-server setup, and Figure~\ref{fig:simulation-results} reports bandwidth and request latency. Table~\ref{tab:PFC} reports the server count before PFC is observed for each window setting. Reducing the window increases the reported count from 63 without software control to 1,218 with a 4\,KB window. The experiment illustrates the effect of limiting outstanding data in this setup; it does not establish a universal PFC-free scale limit.

\begin{table}[t]
 \centering
 \begin{tabular}{lr}
  \toprule
  \textbf{Window size} & \textbf{Servers before PFC} \\
  \midrule
  4\,KB & 1,218 \\
  8\,KB & 567 \\
  16\,KB & 273 \\
  No software control & 63 \\
  \bottomrule
 \end{tabular}
 \caption{Reported server count before PFC is observed, by software window size.}
 \label{tab:PFC}
\end{table}

\section{Maintaining Service Availability}
\label{sec:availability}

\subsection{Transport Reliability \& Service Recovery}
\label{subsec:avail_sys}

Online inference must return results within service latency targets, while training must make progress despite failures. Fan-out makes both sensitive to slow sub-requests: a single stalled destination can delay a larger operation. RC handles transport-level delivery on a functioning connection, but recovery from node failures, persistent path failures, and unavailable replicas requires decisions above the transport layer.

Network monitoring can help locate faults~\cite{guo2015pingmesh,arzani2018007}, and prior deployments combine network and application mechanisms for recovery~\cite{gao2021cloud}. We follow the same general principle: retain service-level recovery in Archon and add the RDMA-specific monitoring and fallback mechanisms needed by our deployment.

\subsection{Deployment and Recovery Mechanisms}
\label{subsec:practical-design}

Archon supported production inference and training before the RDMA deployment, including RPC-level availability policies and load balancing. Integrating \beshort preserves those policies. Before deployment, we test RDMA-equipped machines with Collie~\cite{xinhao2022collie} to identify performance anomalies. We also use a watchdog-based approach to contain PFC storms~\cite{guo2016rdma}.

\paragraphb{Replication across failure domains.}
Rankers and parameter servers are deployed across multiple geographic data centers. Within a data center, parameter-server replicas are placed in different mini-pods, each replica containing the $n$ shards described in $\S$\ref{sec:background:system}. The network between mini-pods within a big-pod supports the lossless RDMA configuration. Replication provides alternative service destinations when a node or path fails; integrated network monitoring helps locate the underlying problem.

\paragraphb{Connection monitoring.}
\beshort detects QP errors through completion polling and verbs return values. It also sends keepalive messages to detect failures such as a remote kernel panic or power loss. Each engine periodically sends its peer a monotonically increasing sequence number. If the peer's sequence number does not advance before the keepalive timeout, the engine enters an error state, notifies the upper layer, and reclaims connection resources. The send interval and failure-detection timeout must accommodate normal scheduling delays.

\paragraphb{RPC-assisted recovery.}
Archon uses separate connection and request timeouts, chosen for the cluster topology and service requirements. Example thresholds are $T_c=10$\,ms for connections and $T_r=40$\,ms for requests. A timed-out request may be retried probabilistically to tolerate transient failures. These transport and request timeouts serve different purposes from the keepalive mechanism.

Recovery proceeds through several options. First, the system attempts to change the connection path using the RNIC's port-affinity support~\cite{nvidia2021mlnx}; the reported deployment uses MLNX OFED 5.0 or later with compatible firmware. If reconnecting does not resolve the failure, Archon selects a nearby replica in another mini-pod. If RDMA communication remains unavailable, it falls back to TCP, following the general approach used in prior RDMA deployments~\cite{gao2021cloud}. In the reported deployment, this fallback restores high throughput within one minute. Section~\ref{sec:lessons} discusses the need to characterize recovery more systematically.

Archon also temporarily quarantines destinations with repeated failures. For example, within a five-minute window, more than $M=30$ connection failures or $N=50$ request failures to a node can place it on an exclusion list. Other replicas serve requests while engineers investigate the node offline. The effectiveness of this policy depends on replica health and available capacity.

These mechanisms describe the recovery architecture, not an end-to-end availability guarantee. In particular, retry safety for state-changing operations such as \texttt{Push} depends on RPC and application semantics. The measurements in this paper do not establish exactly-once execution or quantify availability under all failure modes.

\begin{figure*}[t]
    \centering
    \includegraphics[width=0.92\textwidth]{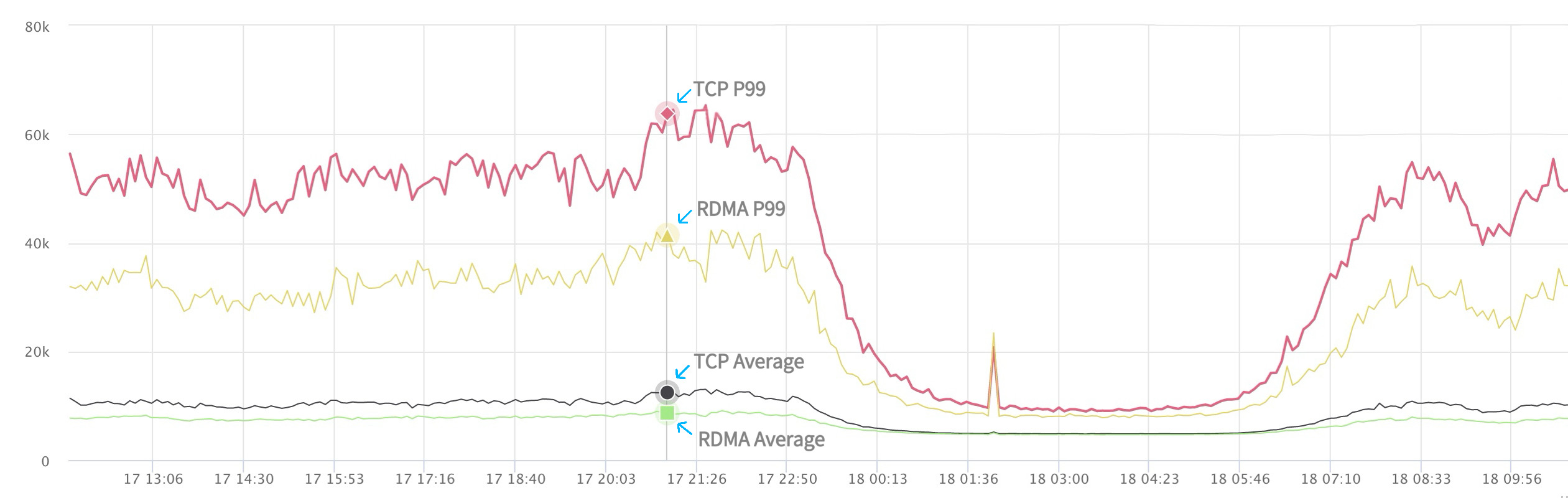}
    \vspace{-1mm}
    \caption{PS performance in a production recommendation inference workload: mean and P99 for TCP and RDMA in microseconds, sampled at one-minute intervals for the busiest rankers.}
    \vspace{-1mm}
    \label{fig:ps-lat-toutiao}
\end{figure*}

\section{Production Deployment and Evaluation}
\label{sec:deployment}

At the time of this study, \bytedance had deployed the RDMA-enabled inference and training systems in tens of data-center clusters over a period of more than two years, supporting services including Toutiao, TikTok, and Xigua Video. This section reports CPU utilization and request latency from two production workloads.

Our baseline is the deployed Linux kernel TCP/IP stack. It provides an operational reference for the migration, but the available results do not compare all possible TCP tuning choices or alternative user-space transports. The microbenchmarks in $\S$\ref{subsec:perf-eval} provide separate comparisons with eRPC and RDMA test baselines.

\subsection{Online Inference}

\begin{table}[t]
 \centering
 \begin{tabular}{lrrr}
  \toprule
  \textbf{CPU metric} & \textbf{RDMA} & \textbf{TCP} & \textbf{Reduction} \\
  \midrule
  Total CPU usage & 51.31\% & 57.45\% & 10.69\% \\
  Software interrupts & 0.16\% & 1.70\% & 90.6\% \\
  System & 4.18\% & 7.08\% & 41.0\% \\
  User & 47.0\% & 48.7\% & 3.5\% \\
  \bottomrule
 \end{tabular}
 \caption{CPU usage during the peak-load Toutiao inference interval. Reductions are relative to TCP.}
 \label{tab:cpu_usage_online}
\end{table}

Figure~\ref{fig:ps-lat-toutiao} shows a production dashboard for a Toutiao recommendation service. The cluster contains 300 parameter servers and 124 rankers, each equipped with a 25\,Gbps Mellanox ConnectX-5 NIC. We measure mean and 99th-percentile (P99) latency for parameter \texttt{Fetch} requests. The RDMA and TCP measurements are reported at matched operation loads.

\begin{table}[t]
 \centering
 \begin{tabular}{lrrr}
  \toprule
  \textbf{Latency} & \textbf{RDMA} & \textbf{TCP} & \textbf{Reduction} \\
  \midrule
  Mean & 8.1\,ms & 12.6\,ms & 35.7\% \\
  P99 & 35.5\,ms & 63.1\,ms & 43.7\% \\
  \bottomrule
 \end{tabular}
 \caption{PS Pull latency at inference peak.}
 \label{tab:latency_online}
\end{table}

The relative latency gap is largest during peak load. Tables~\ref{tab:cpu_usage_online} and~\ref{tab:latency_online} summarize the 21:00--22:00 interval. RDMA reduces total CPU usage from 57.45\% to 51.31\%, a decrease of 6.14 percentage points, or 10.69\% relative to TCP. The largest relative reductions are in software-interrupt and system CPU time, consistent with reducing kernel networking overhead. The results do not separately quantify how the saved CPU capacity is used by the application.

Mean fetch latency decreases by 4.5\,ms (35.7\%), and P99 latency decreases by 27.6\,ms (43.7\%). These are parameter-fetch measurements, rather than end-to-end recommendation latency. They show that reducing communication overhead benefits the measured service, while leaving open how much of the gap a differently configured TCP baseline could recover.

\subsection{Offline Training}

\begin{table}[t]
 \centering
 \small
 \begin{tabular}{lrrr}
  \toprule
  \textbf{Metric} & \textbf{TCP} & \makecell{\textbf{RDMA}\\\textbf{HW only}} & \makecell{\textbf{RDMA}\\\textbf{HW + SW}} \\
  \midrule
  \texttt{Fetch} P99 & 650\,ms & 1.04\,s & 338\,ms \\
  \texttt{Push} P99 & 1.03\,s & 1.12\,s & 268\,ms \\
  PFC duration & N/A & 1\,s per 30\,s & 0 \\
  Loss indicator & 1\%--4\% & \makecell{3K NAKs per 30\,s} & 0 \\
  \bottomrule
 \end{tabular}
 \caption{Metrics for the busiest trainers. HW: hardware congestion control; SW: software send window. TCP loss rates and RDMA NAK counts are not directly comparable.}
 \label{tab:gain_offline}
\end{table}

The training evaluation runs an internal production job on TCP and RDMA clusters with 124 parameter servers and 346 trainers, each equipped with a 25\,Gbps Mellanox ConnectX-5 NIC. Trainers repeatedly issue \texttt{Fetch} and \texttt{Push} requests to retrieve parameters and submit updates.

Table~\ref{tab:gain_offline} distinguishes RDMA with hardware congestion control alone from RDMA with the additional \beshort software window. Both RDMA configurations retain DCQCN and the fabric's PFC configuration. Without the software window, the measured P99 fetch and push latencies exceed those of TCP, and the deployment observes PFC pauses. With the window, P99 fetch latency falls from 1.04\,s to 338\,ms. P99 push latency is 268\,ms, a 74.0\% reduction relative to TCP's 1.03\,s. No PFC pauses or negative acknowledgments are reported during the observation interval. These measurements support bounding outstanding data for this workload; they do not show that the fabric is congestion-free or that PFC can be disabled safely.

\section{Deployment Lessons in Production}
\label{sec:lessons}

Our experience deploying RDMA in production ML inference and training systems yielded several practical lessons.

\paragraphb{Integrate RDMA buffer ownership across layers.}
The practical value of \bemalloc is the buffer-ownership contract it enables across the application and transport. Registered memory becomes useful to applications when allocation, serialization, transfer, and reclamation follow a consistent lifetime model. Integrating this contract beneath Archon's RPC interface allows applications to benefit from zero-copy transfers while retaining their existing communication abstraction.

\paragraphb{Control receiver capacity and outstanding traffic separately.}
Receive credits protect posted receive capacity, while the software send window bounds outstanding data above DCQCN and PFC. In the production training workload, hardware congestion control alone left P99 latency above the deployed TCP baseline. Adding the software window reduced P99 push latency to 268\,ms, with no PFC pauses observed during the measurement interval (Table~\ref{tab:gain_offline}). The operational lesson is to assign each control a specific resource to protect.

\paragraphb{Align RDMA resource sharing with IO thread ownership.}
Assigning each engine to one I/O thread removed contention in our own data structures, but provider-side spin locks still limited message rate as QP counts grew. Thread domains aligned provider resources with application ownership, while per-thread queue sharing reduced resource overhead ($\S$\ref{subsec:multi-thread}). Effective scaling therefore required inspecting contention across the application--provider boundary.

\paragraphb{Keep RDMA transport reliability close to service policy.}
Archon retained responsibility for request timeouts, replica selection, and TCP fallback, while \beshort supplied connection-error and keepalive signals ($\S$\ref{sec:availability}). This division lets recovery use service context, including replica placement across failure domains and the exclusion of repeatedly failing destinations. Preserving these policies was a central part of integrating RDMA into an existing production service.

\section{Related Work}
\label{sec:related-work}

\paragraphb{RDMA systems and machine learning.}
RDMA has been used in key-value stores~\cite{kalia2014using,lim2014mica}, distributed transactions~\cite{wei2015fast,dragojevic2015no}, distributed storage systems~\cite{huang2019bor,huang2019rdma}, cloud security system~\cite{liu2023remote}, disaggregation~\cite{shan2017distributed,huang2025towards,mei2026distdpu}, and RPC~\cite{kalia2019datacenter,tsai2017lite}. EFLOPS~\cite{dong2020eflops} studies RDMA-enabled distributed training through system and algorithm co-design, while Xue \etal~\cite{xue2019fast} integrate RDMA into distributed machine learning. 
Rather, our focus is the operational RDMA co-design across registered-memory allocation, message transport, resource sharing, and RPC-level recovery for PS-based inference and training.

\paragraphb{Connection scaling.}
QP context and resource sharing are established RDMA design concerns~\cite{kalia2016design,monga2021birds}.
SkyRDMA~\cite{huang2027skyrdma} explores offloaded RDMA connection sharing policies at the hardware level in AI clouds.
Prior work also explores software-defined scheduling of RNIC resources~\cite{huang2025exposing} and GPU fabrics~\cite{chen2026unlocking}.
Dynamically Connected transport provides another approach: an initiator can communicate with multiple targets without a dedicated RC connection to each~\cite{nvidia_dc}. \beshort instead multiplexes logical sockets over RC QPs and uses thread domains to align provider resources with I/O-thread ownership. 

\paragraphb{Congestion control.}
DCQCN~\cite{zhu2015congestion}, TIMELY~\cite{mittal2015timely}, Swift~\cite{kumar2020swift}, and HPCC~\cite{li2019hpcc} explore different congestion signals and transport-control designs. 
Instead, our software window provides application-level control over outstanding RC traffic, layered above the existing DCQCN configuration and complementary to prior studies.

\section{Conclusion}
\label{sec:conclusion}

Deploying RDMA in production inference and training systems requires coordinating application interfaces, memory ownership, transport resources, and service recovery. \belong combines an RDMA-aware allocator and zero-copy transfers with QP multiplexing, thread-aware resource management, receive credits, and a software send window. Archon retains the RPC-level context needed for retries, replica selection, and TCP fallback. The reported deployments demonstrate lower parameter-fetch latency and CPU usage relative to the deployed TCP baseline, while exposing limits that hardware offload alone does not resolve. The broader lesson is to assign each layer a clear responsibility and evaluate the resulting communication path under the concurrency and failure patterns of the target workload.

\begingroup
\small
\bibliographystyle{plain}
\bibliography{rdma}
\endgroup
\end{document}